\def\year{2022}\relax
\documentclass[letterpaper]{article}
\usepackage{aaai22} 
\usepackage{times} 
\usepackage{helvet} 
\usepackage{courier} 
\usepackage[hyphens]{url}  
\usepackage{graphicx}
\usepackage{natbib}  
\usepackage{caption} 
\DeclareCaptionStyle{ruled}{labelfont=normalfont,labelsep=colon,strut=off} 
\usepackage{amsmath}
\usepackage{amssymb}
\usepackage{pifont}
\usepackage{multirow}
\usepackage{threeparttable}
\usepackage{booktabs}
\usepackage{multirow}
 \usepackage{verbatim}
\usepackage[tight,footnotesize]{subfigure}

\usepackage{algorithm}
\usepackage{algorithmic}
\usepackage{newfloat}
\usepackage{listings}
\floatstyle{ruled}
\newfloat{listing}{tb}{lst}{}
\floatname{listing}{Listing}
\title{Joint Audio-Text Model for Expressive Speech-Driven 3D Facial Animation}

\author {
   Yingruo Fan\textsuperscript{\rm 1},
    Zhaojiang Lin\textsuperscript{\rm 2},
    Jun Saito\textsuperscript{\rm 3},
    Wenping Wang\textsuperscript{\rm 1,4},
    Taku Komura\textsuperscript{\rm 1}
}
\affiliations {
    \textsuperscript{\rm 1}The University of Hong Kong\\
    \textsuperscript{\rm 2}The Hong Kong University of Science and Technology\\
    \textsuperscript{\rm 3}Adobe Research\\
    \textsuperscript{\rm 4}Texas A\&M University\\
    \{yrfan,wenping,taku\}@cs.hku.hk, zlinao@ust.hk, jsaito@adobe.com
}
\usepackage{bibentry}

\begin{document}

\maketitle

\begin{abstract}
Speech-driven 3D facial animation with accurate lip synchronization has been widely studied. However, synthesizing realistic motions for the entire face during speech has rarely been explored. In this work, we present a joint audio-text model to capture the contextual information for expressive speech-driven 3D facial animation. The existing datasets are collected to cover as many different phonemes as possible instead of sentences, thus limiting the capability of the audio-based model to learn more diverse contexts. To address this, we propose to leverage the contextual text embeddings extracted from the powerful pre-trained language model that has learned rich contextual representations from large-scale text data. Our hypothesis is that the text features can disambiguate the variations in upper face expressions, which are not strongly correlated with the audio. In contrast to prior approaches which learn phoneme-level features from the text, we investigate the high-level contextual text features for speech-driven 3D facial animation. We show that the combined acoustic and textual modalities can synthesize realistic facial expressions while maintaining audio-lip synchronization. We conduct the quantitative and qualitative evaluations as well as the perceptual user study. The results demonstrate the superior performance of our model against existing state-of-the-art approaches.
\end{abstract}

\section{Introduction}

\begin{figure}
\centering
\includegraphics[width=0.5\textwidth]{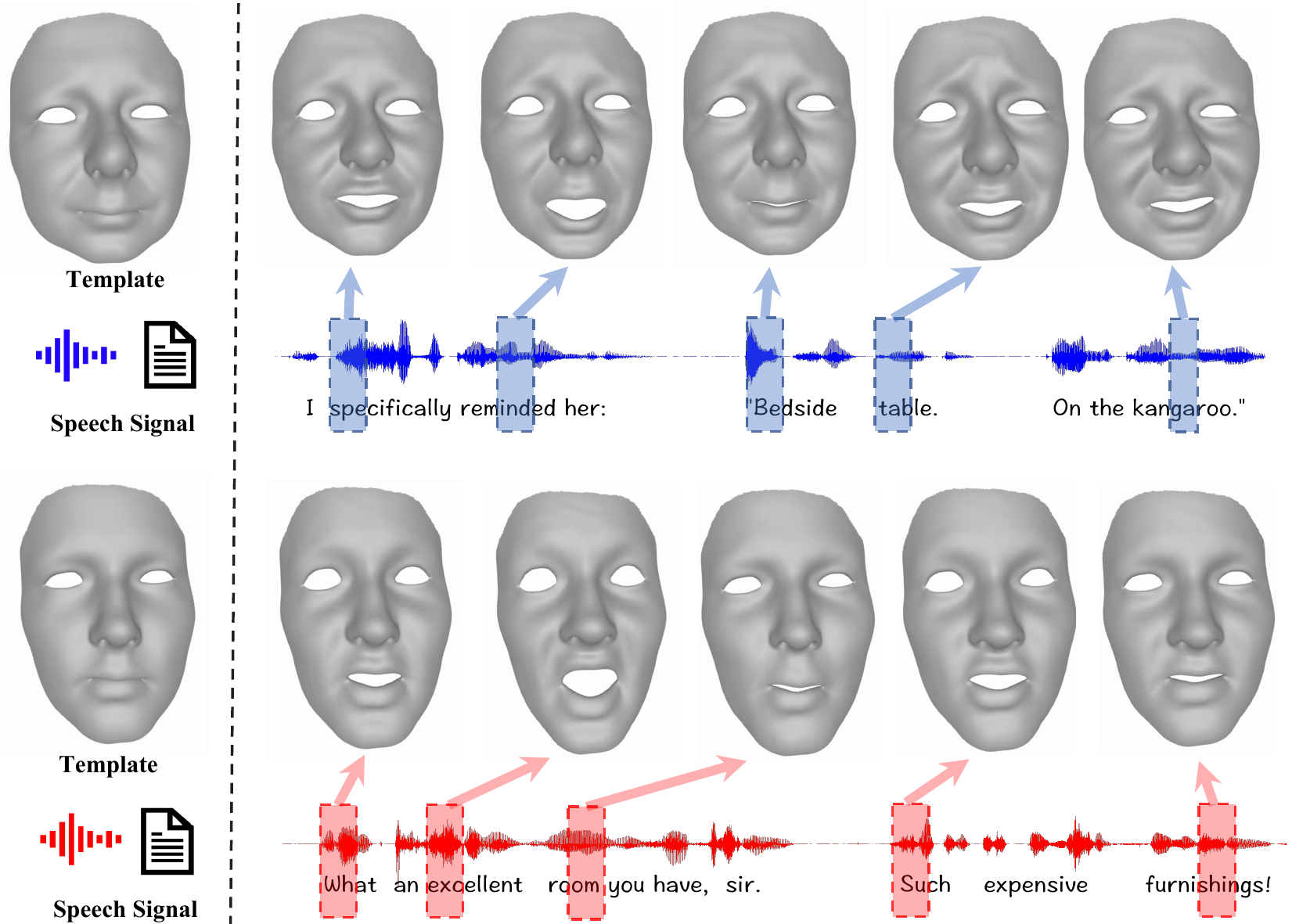}
\caption{\label{fig:brief}
Given a speech signal (audio and text) and a 3D face mesh, our method produces the expressive lip-synchronized 3D facial animation with realistic facial expressions. Please see our supplementary video.}
\end{figure}

Speech-driven 3D facial animation has garnered tremendous interest in computer graphics and vision. It can be used in a wide array of potential applications, including computer games, filmmaking, 3D telepresence system and other human-computer interactive interfaces. Generating expressive lip-synchronized 3D facial animation from speech remains challenging. There appear to be several reasons for this. First, the subtle changes in upper face expressions are weakly correlated with the audio~\cite{cudeiro2019capture,karras2017audio}. The same audio signal can be associated with different upper facial motion sequences and vice versa. Second, synthesizing natural facial muscle movements is difficult due to the complicated geometric structure of human faces~\cite{edwards2016jali}. Third, the acoustic speech signal can vary from person to person due to the identity and physical differences between speakers~\cite{sjerps2019speaker}.

Most existing studies on audio-driven 3D facial animation~\cite{zhou2018visemenet,edwards2016jali,taylor2017deep,suwajanakorn2017synthesizing} focus on the issue of lip-synchronization, which might lack realism in the upper face such as eyelids and brows. However, it is important to animate a 3D talking avatar that produces speech utterances with vivid facial expressions. A small collection of studies~\cite{karras2017audio,richard2021meshtalk,wang20213d} emphasize on the synthesized facial motions of the entire face as well as the lip-sync. The latest state-of-the-art approach~\cite{richard2021meshtalk} adopts a cross-modality loss to disentangle the audio-correlated and audio-uncorrelated face motions, thus ensuring the plausible animation of the upper face. Similar to~\cite{karras2017audio,richard2021meshtalk}, our focus is animating a 3D face mesh with realistic muscle movements in both the upper and lower face parts. On the other hand, we aim to predict the geometry of a 3D talking head instead of the pixel values of a 2D talking head image. In this paper, we present a novel approach that combines acoustic and textual modalities to animate a 3D talking head with vivid facial expressions, as illustrated in Figure~\ref{fig:brief}. Given the audio and text signal, our method generates expressive 3D face animation sequences which have synchronized lip movements to the audio and realistic facial muscle movements (e.g., eye and eyebrow motions) for the entire face. 

The human speech signal inherently involves both acoustic and textual features. Emotions can manifest acoustically in measures such as the speaking rate and the fundamental frequency. On the other hand, text contents can also convey the emotional state of the speaker. For example, if the speaker frowns and says ``It is the most dreadful news'' in a neutral voice, then it would be ambiguous when inferring the facial expression from the acoustic features. In this case, the emotional state is mostly reflected through the expression of language, e.g., the word ``dreadful''. Inspired by this, we investigate the hypothesis that integrating acoustic and textual context could improve speech-driven 3D facial animation.

The pre-trained transformer-based representations~\cite{devlin2018bert,radford2019language} have proven to be successful in various natural language processing tasks. Our motivation is that the pre-trained language model has learned rich contextual information, since it has been trained on the large scale text corpora. While for the existing 3D audio-visual datasets of facial animation~\cite{fanelli2010,richard2021meshtalk,karras2017audio}, they only have a limited number of sentences. Training a model with audio alone might limit its ability to learn more diverse contexts. Intuitively, the high-level textual features could aid in understanding the emotional context of the speech. Previous studies on speech-driven gesture generation~\cite{yoon2020speech,kucherenko2020gesticulator} have demonstrated the effect of the text modality and verified that using text and audio modalities together can further improve their results. Therefore, we exploit the text
embeddings from GPT-2~\cite{radford2019language} and incorporate them as part of input features to our model. This is different from previous linguistics-based methods which usually extract the phoneme-level features from the transcript. To the best of our knowledge, there has been no previous attempt at exploring the language model to resolve the ambiguity of facial expression variations for speech-driven 3D facial animation.

In our method, the language and acoustic modalities are modeled through two subnetworks separately. Furthermore, it is desirable to take into account the bimodal interactions between audio and text cues. To this end, we build a fusion layer, named Tensor Fusion~\cite{zadeh2017tensor}, in our model to disentangle unimodal and bimodal dynamics. The Tensor Fusion layer explicitly models the unimodal and bimodal interactions based on the Cartesian product of audio and text embeddings. Ideally, considering audio and text modalities and their interactions should capture a wide range of variations in the speech. This allows the model to learn more expressive joint representations for facial expressions.

In summary, our main contributions include:

\begin{itemize}
    \item Exploiting the contextual text embeddings for speech-driven 3D facial animation. Considering the potential use of a strong pre-trained language model, this work has implications for producing more expressive facial motions using audio and text signals as inputs.
    
    \item Providing meaningful insights into the effect of the text modality on the upper part of the predicted face mesh by performing visualization analysis.
    
    \item Extensive evaluation of the proposed approach on a 3D audio-visual corpus~\cite{fanelli2010} and a perceptual user study of facial animation quality in terms of realism. The results show that our approach can synthesize more realistic 3D facial animations, as compared to existing state-of-the-art approaches.
    
\end{itemize}

\begin{figure*}
\centering
\includegraphics[width=0.9\textwidth]{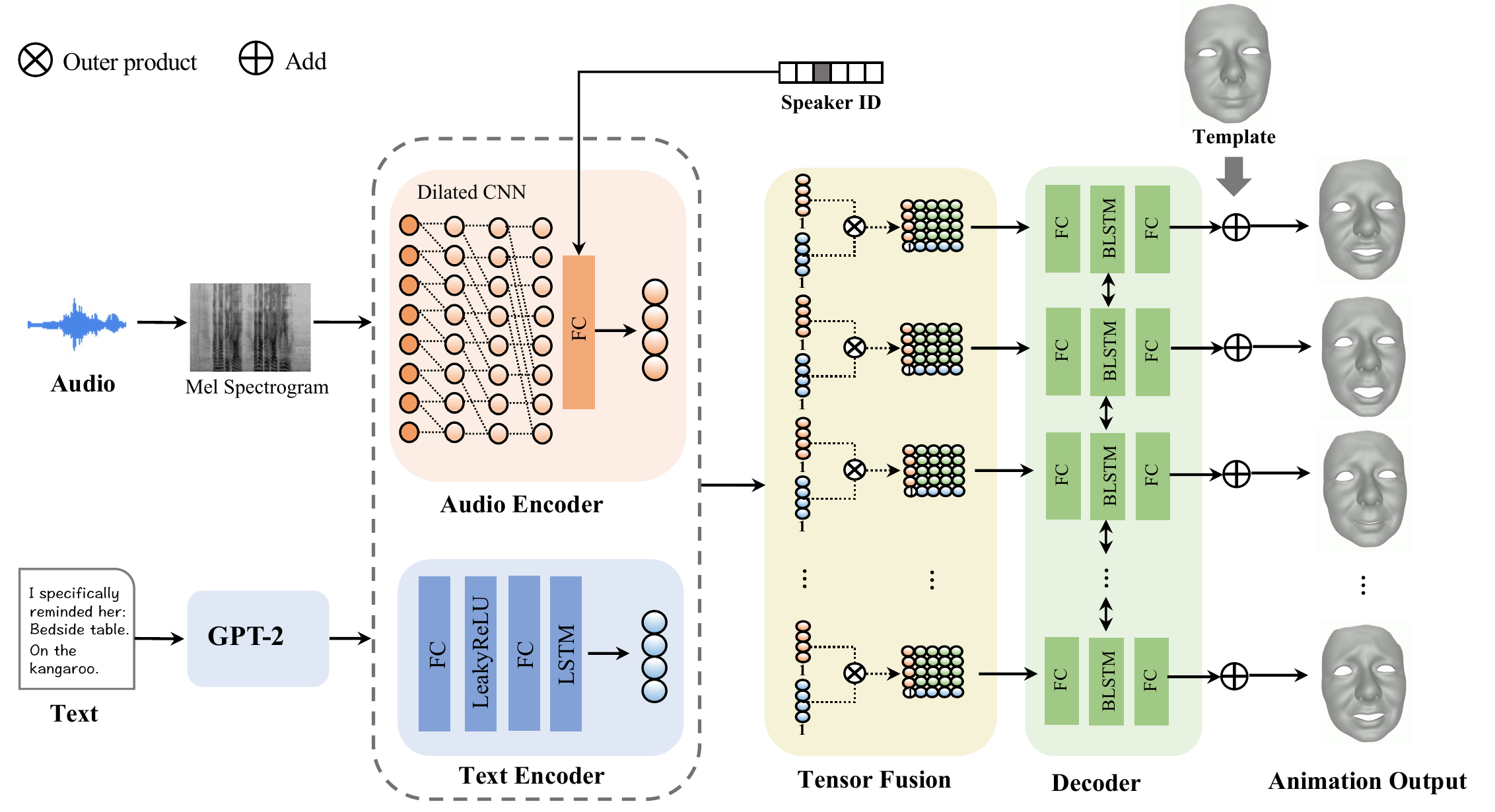}
\caption{\label{fig:framework}Overview of our method. First, the raw audio is converted into mel spectrogram whereas the text representations are extracted from the pre-trained GPT-2. Second, these extracted features are fed to the respective encoders to produce the audio and text embeddings. The one-hot speaker embedding is concatenated to the output of the final convolution layer of the audio encoder. Third, we employ a tensor fusion layer to fuse the encoded contextual features of audio and text modalities. Finally, the decoder produces a sequence of animated 3D face meshes, each of which is represented as a 23370-dimensional vector in 3D vertex coordinates. }
\end{figure*}

\section{Related Work}

Speech-driven facial animation has been a popular topic in computer graphics and vision. In computer vision, it is also known as talking face generation. Here we mainly review the previous speech-driven 3D  facial animation approaches, where the output is the 3D mesh animation. The related works can be categorized into two broad groups: linguistics-based methods and audio-based methods.

\subsection{Linguistics-based Methods}

The concept of phonemes is well developed in speech animation. Viseme~\cite{fisher1968confusions} is derived from a set of phonemes that have similar visual appearances on the lips.
The popular dominance model~\cite{massaro2012} uses the dominance functions of consecutive phonemes and determines the influence of each phoneme on the respective facial animation parameters. Taylor et al.~\shortcite{taylor2012dynamic} propose the one-to-many phoneme to viseme mappings, a.k.a. dynamic visemes, using a data-driven method. Xu et al.~\shortcite{xu2013practical} define the animation curves for a canonical set of visemes to achieve the coarticulation effects. Alternatively, Edwards et al.~\shortcite{edwards2016jali} introduce the linguistic rules to animate the phonemes of a JALI (Jaw and Lip) rig, which is built upon the Facial Action Coding System (FACS). The follow-up work VisemeNet~\cite{zhou2018visemenet} predicts a stream of phoneme-groups and facial landmarks from audio, and produces the JALI parameters used for animating the 3D face.

Previous linguistics-based approaches focus on synthesizing the synced mouth movements and explore diverse forms of complex linguistic rules, including phoneme-to-viseme mapping and coarticulation. Unlike them, we aim to discover high-level semantic features that might
correspond to the emotional state, as such text representations would be useful for facial expression synthesis.

\subsection{Audio-based Methods}

Recently, a variety of approaches for animating the 3D talking avatar from audio have been proposed. The earlier method~\cite{cao2005expressive} implements a support vector machine classifier to detect the emotional state of the input audio and then synthesizes the 3D animations. Pre-labeling emotion categories is required, such as happy, sad and angry, for all the sentences. Karras et al.~\shortcite{karras2017audio} disambiguate the variations in facial expression by introducing a latent emotional state vector. They collect 3–5 minutes of animation data for two subjects. However, their model is person-specific and not applicable to unseen speakers. Taylor et al.~\shortcite{taylor2017deep} propose a sliding window approach on phoneme subsequences and generalize their model to other avatars via re-targeting. Pham, Cheung, and Pavlovic~\shortcite{pham2017speech} obtain the blendshape coefficients from the 2D videos first. Subsequently, the predicted coefficients are regarded as ground truth for training an audio-to-face model. Despite the generalization ability, both methods~\cite{taylor2017deep,pham2017speech} rely on 2D videos rather than high-resolution 3D face scans, which may influence the quality of the resulting animation. The CNN-based model VOCA~\cite{cudeiro2019capture} extracts the audio features from a pre-trained DeepSpeech model~\cite{hannun2014deep} and then learns the mapping of audio to the 3D vertex offsets of a face model. However, as can be noted, VOCA only learns the facial motions that are mostly present in the lower part of the face. In a recent work, MeshTalk~\cite{richard2021meshtalk}, the cross-modality loss is employed to disentangle the information about upper and lower face motion, thus generating full facial 3D animation from speech.
 
The approaches closely related to ours are VOCA~\cite{cudeiro2019capture}, MeshTalk~\cite{richard2021meshtalk} and the Audio2Face model~\cite{karras2017audio} in the sense that their models are trained on high-resolution 3D face scans and are designed to decode the learned representations to 3D vertex space. Different from their methods using audio features alone, we propose to incorporate the contextual embeddings from Transformer-based GPT-2 to aid in understanding the emotional context, in order to produce a more diverse range of facial expressions.

\section{Method}

\subsection{Overview of the Architecture}

The pipeline of our proposed method is illustrated in Figure~\ref{fig:framework}. Overall, the network is composed of four main components: the audio encoder, the text encoder, the tensor fusion module and the decoder. The speaker identity is represented as a one-hot vector. The audio-text embeddings are mapped into a high-dimensional space of 3D vertex offsets, i.e., 70110 outputs. In the end, the predicted vertex offsets are added to the vertices of the neutral template mesh, generating the expressive 3D face animation sequence.

\subsection{Problem Formulation}

Let us assume a video sequence of $\mathbf{T}$ face meshes $\mathbf{Y}=[{\mathbf{y}}_{t}]_{t=1:\mathbf{T}}$, each of which has $\mathbf{V}$ vertices in 3D coordinates. Given a sequence of acoustic features $\mathbf{X^{a}}=[{\mathbf{x^{a}}}_{t}]_{t=1:\mathbf{T}}$ as well as the textual features $\mathbf{X^{l}}=[{\mathbf{x^{l}}}_{t}]_{t=1:\mathbf{T}}$, our goal is to predict the corresponding 3D facial animation sequence $\mathbf{\tilde{Y}}=[{\mathbf{\tilde{y}}}_{t}]_{t=1:\mathbf{T}}$. Since the speech signal and the facial movements are synchronized temporally, we explicitly align the audio and text features to the corresponding video frame.

\subsection{Audio Encoding}

Inspired by~\cite{oord2016wavenet,richard2021meshtalk}, we adopt a stack of four dilated temporal convolutional layers as the first part of the audio encoder. Dilated convolutions~\cite{yu2015multi} were first proposed in computer vision for context aggregation. Here we use the time-dilated convolution so that the feature maps and output can keep the high-resolution information of the input in the temporal domain. Formally the dilated discrete convolution operation can be described as follows:

\begin{equation}\label{eq-1}
\left(\mathbf{F} *_{l} \mathbf{k}\right)(\mathbf{p})=\sum_{s+lt=\mathbf{p}} \mathbf{F}(s) \mathbf{k}(t), t \in[-r, r],
\end{equation}
where $*_{l}$ denotes the convolution with dilation factor $l$,  $\mathbf{F}$ is a discrete function, $\mathbf{p}$ is the feature map, and $\mathbf{k}$ is the filter with size $(2 r+1)^{2}$. Particularly, $*_{l}$ supports exponentially increasing the receptive field without losing resolution. Hence, we empirically use a dilation factor that is exponentially increased with respect to the network depth. Each convolution layer has the same number of filters and the same kernel size, followed by the LeakyReLU activation. Besides, the residual connections are used to maintain the temporal information about input acoustic features.

 Similar to~\cite{cudeiro2019capture}, we indicate the speaker identity with a one-hot vector to model the speaking style. This one-hot vector is concatenated to the output of the final convolution layer of the audio encoder, forming the input of the following fully-connected layer. During inference, altering the one-hot vector can manipulate the output animation in different speaking styles.

\subsection{Text Encoding}
The introduction of transformers~\cite{vaswani2017attention} has shown promise in the field of language and vision. Transformer-based models have been applied to capture the contextual information of different modalities such as text and image~\cite{ramesh2021zero}. To incorporate the contextualized semantic features, we adopt the pre-trained Transformer-based language model, GPT-2~\cite{radford2019language}, to process the input text modality. Specifically, each sentence is encoded into 768 features per word. Due to different sequence lengths of text and audio, we upsample the GPT-2 features to match the length of the audio feature sequence according to the duration time of each word. The text encoder consists of two fully-connected layers, a LeakyReLU layer and an LSTM layer. The outputs of the text encoder are taken as the features for the text modality.  

\subsection{Tensor Fusion}

At the fusion stage, we apply a tensor fusion layer~\cite{zadeh2017tensor} to disentangle the unimodal dynamics and bimodal interactions. Prior studies on multimodal fusion~\cite{zadeh2017tensor,poria2016convolutional} have demonstrated that this method can better capture multimodal interactions than the simple concatenation operation. The tensor fusion layer first obtains the unimodal feature vectors, namely the audio embedding $\mathbf{H^{a}}$ and the language embedding $\mathbf{H^{l}}$, by passing the unimodal inputs $\mathbf{X^{a}}$ and $\mathbf{X^{l}}$ into the audio encoder and the text encoder, respectively. Subsequently, the tensor fusion layer produces the multimodal output representation $\mathbf{H^{m}}$ by performing a differentiable outer product operation, which is computed by:

\begin{equation}\label{eq-2}
\mathbf{H}^{m}=\left[\begin{array}{c}
\mathbf{H}^{a} \\
1
\end{array}\right] \otimes\left[\begin{array}{c}
\mathbf{H}^{l} \\
1
\end{array}\right],
\end{equation}
where $\otimes$ denotes the outer product. The interpretation of tensor fusion is illustrated in Figure~\ref{fig:framework}. The multimodal dynamics are generated by adding an extra constant dimension with value 1. This results in a 2-D Cartesian space, where the horizontal and vertical axes are defined by $[\mathbf{H}^{a} \;1]^{T}$ and $[\mathbf{H}^{l}\;1]^{T}$, respectively.

\subsection{3D Facial Animation Generation}
In the decoder module, the sequence of $\mathbf{T}$ fused features $\mathbf{{H}^{m}}=[{\mathbf{h^{m}}}_{t}]_{t=1:\mathbf{T}}$ obtained in Equation~\ref{eq-2} is fed to a fully-connected layer to project it to a lower-dimensional output vector space. After that, two bidirectional LSTMs are applied to model temporal dependencies. The bidirectional LSTM makes use of both the preceding and succeeding context information by processing the sequence both forward and backward. Finally, the output sequence of two bidirectional LSTMs is used as the input sequence of a fully-connected layer, which maps the learned representation to 3D vertex space. Therefore, the decoding flow can be described as:
\begin{equation}\label{eq-3}
[{\mathbf{\tilde{y}}}_{t}]_{t=1:\mathbf{T}} = \text{FC}(\text{BLSTM}(\text{FC}([{\mathbf{h^{m}}}_{t}]_{t=1:\mathbf{T}}))).
\end{equation}

During the training phase, we minimize the mean squared difference between the ground-truth vertices $\mathbf{Y}=[{\mathbf{y}}_{t}]_{t=1:\mathbf{T}}$ and the corresponding outputs $\mathbf{\tilde{Y}}=[{\mathbf{\tilde{y}}}_{t}]_{t=1:\mathbf{T}}$ produced by the decoder, which is computed as follows:

\begin{equation}\label{eq-4}
\mathcal{L}=\sum_{t=1}^{\mathbf{T}} \sum_{v=1}^{\mathbf{V}} \left\|\mathbf{\tilde{y}}_{t, v}-\mathbf{y}_{t, v}\right\|^{2}.
\end{equation}

\section{Experiments}

\subsection{Dataset}
We train and evaluate our models using the publicly-available  audio-visual dataset BIWI~\cite{fanelli2010}. This corpus involves 14 native speakers (8 females and 6 males), each of which utters 40 short English sentences. The corresponding text transcriptions are also provided. The 3D geometry data of the performance of the speaker is acquired at 25 fps. All face meshes have the same topology and the same number of vertices (23370). For each speaker, the static face mesh with a neutral expression is provided as the template. Each sentence lasts 4.67s long on average and is recorded twice: with and without emotion. We use the sentences recorded in the emotional context for our experiments and remove the sequences that are not complete in terms of the tracked 3D face scans or the audio files. 

We divide the dataset into training, validation and testing parts. The training set is composed of 192 sentences, pronounced by six subjects (each subject utters 32 sentences). The validation part contains 24 sentences, pronounced by the same six subjects (each subject utters 4 sentences). The testing part includes two sets of test sentences: the first are 4 sentences of the same six subjects in the training set, i.e., 24 sentences (Test set $\mathbf{A}$); the second is a set of 4 sentences of eight subjects not seen in the training set, i.e., 32 sentences (Test set $\mathbf{B}$). No overlap of sentences exists in the training, validation, and test sets, and there is no overlap of subjects presents in the training set and Test set $\mathbf{B}$. Here Test set $\mathbf{A}$ is used for quantitive evaluation whereas Test set $\mathbf{B}$ represents a more challenging set used for testing the generalization ability to unseen subjects.

\subsection{Implementation Details}
\subsubsection{Data pre-processing}
To extract the audio features, we utilize the librosa~\cite{mcfee2015librosa} library to transform the raw audio to 128-channel mel-frequency power spectrograms and convert the power spectrograms to decibel (dB) units. Besides, we synchronize the input audio sequence to the corresponding 3D face animation sequence at 25 fps. Hence, the input audio feature vector for every visual frame has 128 dimensions. For the text features, we extract the word representations from the pre-trained GPT-2 small model~\cite{radford2019language} (12 layers, 768-dimensional hidden state). To ensure that the input audio and text feature sequences share the same time steps, we employ the Gentle forced aligner~\cite{ochshorn2017gentle} to find the start and end timestamps of each word within a sentence. Given the known duration time of each word, we apply zero padding to the frames that do not contain a word as there is usually a pause between words. For each word that contains semantic information, we repeat the extracted word representation according to the word's duration time. The text feature vector for every visual frame has 768 dimensions. Then, we set each text feature vector to the average of its surrounding text feature vectors (previous 8 vectors and future 7 vectors). After pre-processing, both the input audio and text feature sequences have the same length as the 3D face animation sequence.

\begin{table}
\centering
 \caption{\label{tb:compare} Quantitative evaluation of different methods. We report the mean absolute error (MAE) between the ground-truth data and the generated results for AU coding consistency evaluation.
 } 
 \begin{tabular}{lccc}
 \toprule
\multicolumn{2}{l|}{\textbf{Methods}}& \textbf{Upper Face AUs} & \textbf{Lower Face AUs} \\
\midrule 
\multicolumn{2}{l|}{VOCA}& 0.278	&0.229\\
\multicolumn{2}{l|}{MeshTalk}& 0.236&	0.230\\
\multicolumn{2}{l|}{Ours}& 0.221&  0.212 \\
\bottomrule
 \end{tabular} 
\end{table}

\begin{table}
\centering

 \caption{\label{tb:AMT} Perceptual user study results. The number reports the percentage ($\%$) of A/B tests where A is chosen over B. 
 } 
 \resizebox{0.48\textwidth}{!}{
 \begin{tabular}{lcccc}
 \toprule
\multicolumn{2}{l|}{\textbf{Model Pairs}}&\textbf{Full Face} &\textbf{Upper Face} &\textbf{Lips}  \\
\midrule 
\multicolumn{2}{l|}{Ours/VOCA}& $84.20\pm6.21$  & $84.20\pm6.17 $&$84.90\pm6.81$ \\
\multicolumn{2}{l|}{Ours/MeshTalk}&  $71.35\pm3.13$ & $68.92\pm5.00$& $71.88\pm3.65$  \\
\multicolumn{2}{l|}{Ours/GT}& $19.79\pm5.73$ &$24.48\pm9.39$ &$19.27\pm6.14$  \\
\bottomrule
 \end{tabular} 
 }
\end{table}

\subsubsection{Network architectures.}
The overall architecture of the model is shown in Figure~\ref{fig:framework}. The audio encoder first applies 4 dilated convolutional layers each equipped with 128 filters of size $3\times 3$. The dilation factors of these convolutions are 1, 2, 4 and 8, respectively. Each dilated convolutional layer is followed by the LeakyReLU activation. Additionally, the encoder uses residual connections between layers. The output of the dilated convolutions is then concatenated with a 6-dimensional one-hot vector. Finally, the concatenated feature vector goes through a 128-unit FC layer. The text encoder consists of a 128-unit FC layer, a LeakyReLU layer, a 64-unit FC layer and a 64-unit LSTM layer. The tensor fusion layer takes a 128-dimensional audio embedding vector and a 64-dimensional text embedding vector as inputs and produces a high-dimensional feature vector. The subsequent decoder consists of a 128-unit FC layer, two 128-unit bidirectional LSTM layers and a 70110-unit FC layer.

\subsubsection{Training setup.}
All the experiments are implemented using PyTorch. In the training stage, the optimization function is Adam, with a constant learning rate of 1e-4 and a batch size of 1. We train our model for 100 epochs and apply it to the testing data directly.

\subsection{Comparisons Against State-of-the-Arts}

We compare our method with the state-of-the-art approaches in speech-driven 3D facial animation, VOCA~\cite{cudeiro2019capture} and MeshTalk~\cite{richard2021meshtalk}. We use the publicly available implementation for VOCA and train the model on BIWI. Since the implementation of MeshTalk is not publicly available yet, we reproduce their method to the best of our understanding.

\subsubsection{Quantitative Evaluation}

Given the many-to-many mappings between the speech signal and the facial expressions, it is not adequate to use the norm on the prediction error between the original and generated outputs to measure the quality of speech-driven 3D facial animation, as suggested by the previous study~\cite{cudeiro2019capture}. The related works~\cite{cudeiro2019capture,karras2017audio} only include the perceptual and qualitative evaluations. Since we focus on the realism of facial expressions, we perform the quantitative evaluation based on the popular facial action unit representation FACS~\cite{eckman1978facial}.
FACS is the ``gold standard'' for quantifying facial muscle movements, a.k.a., Action Units (AUs). The combination of different AUs can represent all possible facial expressions. Here we first render both the ground-truth and generated 3D geometry data with texture as 2D videos with a resolution of $800\times800$. Then we utilize OpenFace~\cite{baltruvsaitis2016openface} to detect the AU intensities for the rendered videos. The mean absolute error (MAE) is used as the performance metric for evaluating the AU coding consistency between the ground-truth and generated results. We categorize all AUs into two groups: upper face AUs (AU1,2,4,5,6,7,9 and 45) and lower face AUs (AU10,12,14,15,17,20,23,25 and 26), and compute the average MAEs for these two groups, respectively. The quantitative evaluation for all the methods on Test set $\mathbf{A}$ of the BIWI dataset is summarized in Table~\ref{tb:compare}. 

Since the subtle facial movements are difficult to quantify, we recommend watching the clips in the supplementary video to assess the quality of the results.

\subsubsection{Perceptual Evaluation}
We carry out the perceptual user study on Amazon Mechanical Turk (AMT). We compare our approach to VOCA, MeshTalk and the captured ground-truth data using all sentences of Test set $\mathbf{B}$, i.e., 32 sentences each spoken by a subject not seen in the training set. For VOCA and our method, we obtain the prediction results conditioned on all six training identities. Therefore, 192 (32 sentences $\times$ 6 identities) pairs are evaluated for each row in Table~\ref{tb:AMT}. In total, 576 A vs. B pairs are created. In our study, a HIT consists of four pairs, one of which is used for qualification test (ground-truth vs. result produced by a not well trained model). Each HIT is evaluated by 3 judges and a total of 576 HITs are collected. Turkers are required to make the right choice for the qualification test before they are allowed to submit HITs. For each pair, two animations with the same subject and audio are presented side-by-side and three questions are designed in terms of the full face, the upper face and the lips. Turkers are instructed to watch the videos and asked to determine which animation looks more realistic for each given question. We shuffle the order of the videos and also randomize which method is A and which is B for each pair to eliminate any bias. The user interface and more details about the user study can be found in Supplementary Material.

Table~\ref{tb:AMT} shows the comparison of success rates of our method against VOCA, MeshTalk, and the ground truth. In comparison to VOCA, our method wins about 84$\%$ of the pairwise comparisons for the three face parts. This is not surprising as VOCA does not synthesize the upper face movements well. Turkers perceive the animation results of our method more realistic than those of MeshTalk, with the success rate of 71.35$\%$ for the full face, 68.92$\%$ for the upper face and 71.88$\%$ for the lips. As expected, the animations synthesized by our approach are generally perceived as less realistic than the ground truth. The result is similar to that reported in~\cite{cudeiro2019capture}, as inferring from speech alone can not synthesize the subtle subject-specific details of the unseen subjects~\cite{cudeiro2019capture}.

\begin{table}
\centering
 \caption{\label{tb:ablation} Ablation experiments on different input modalities and different fusion methods. The evaluation metric is mean absolute error. \textit{\textbf{C}} = Concatenate, \textit{\textbf{TF}} = Tensor Fusion.
 } 
 \resizebox{0.48\textwidth}{!}{
 \begin{tabular}{lccc}
 \toprule
\multicolumn{2}{l|}{\textbf{Methods}}& \textbf{Upper Face AUs} & \textbf{Lower Face AUs}  \\
\midrule 
\multicolumn{2}{l|}{Audio Only}& 0.251 & 0.224 \\
\multicolumn{2}{l|}{Text Only}& 0.247 & 0.253 \\
\multicolumn{2}{l|}{Audio+Text (\textit{\textbf{C}})}& 	0.225 & 0.218 \\
\multicolumn{2}{l|}{Audio+Text (\textit{\textbf{TF}})}& 0.221&  0.212 \\
\bottomrule
 \end{tabular} 
 }
\end{table}

\subsubsection{Qualitative Evaluation}

The supplementary video includes qualitative comparisons with the captured performance data, VOCA and MeshTalk. We also compare our results with the dynamic viseme method~\cite{taylor2017deep} and the emotion-driven method~\cite{karras2017audio} using the footage from their supplementary videos. To validate the ability to synthesize different speaking styles for the same speech, we provide the generated results by conditioning on all six training subjects. To assess the generalization capability of our model, we test our model using the speech clips extracted from TED videos on YouTube. As shown in the supplementary video, our method produces realistic and natural-looking facial animations, synthesizes various speaking styles and generalizes to unseen voices. More importantly, the generated facial motions appear to be expressive when the emotional state of the speaker is obvious. We can notice that the talking avatar shows more expressive facial movements when saying certain emotional words, e.g., ``great'', ``thank you'', etc. Please refer to the supplementary video for qualitative results.

\begin{figure}
\centering
\includegraphics[width=0.35\textwidth]{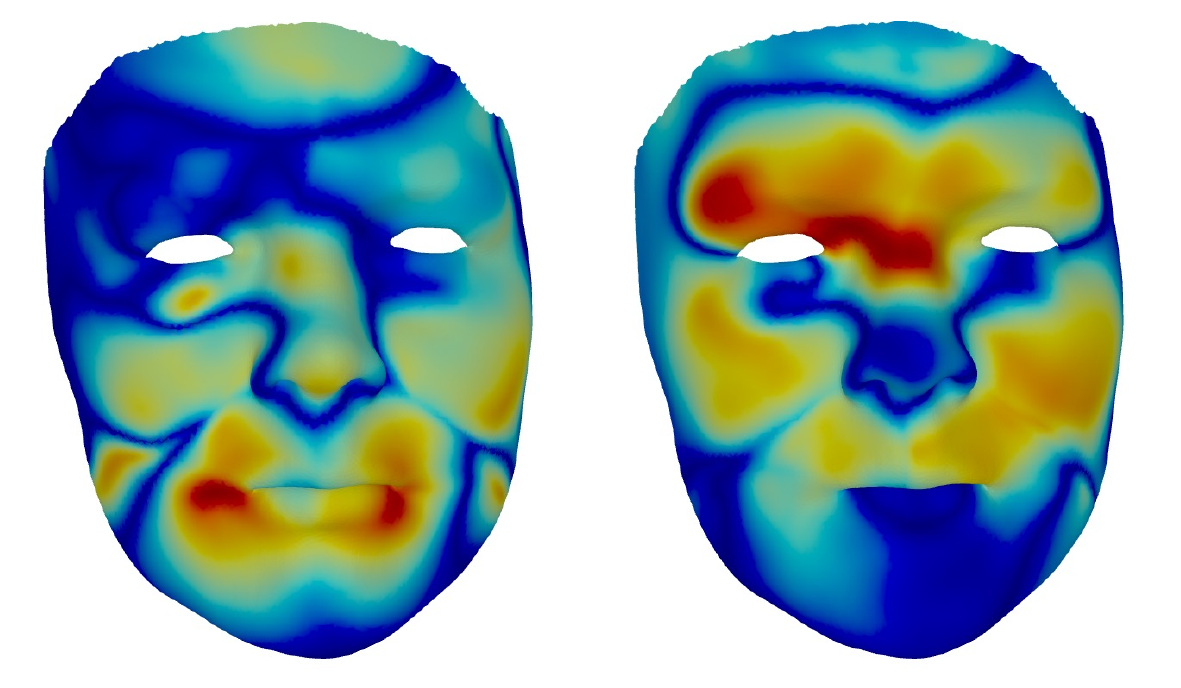}
\caption{\label{fig:correlation}
Correlation between the encoded features (left: audio; right: text) and the predicted vertex offsets. The correlation is measured by the absolute value of the Pearson Correlation Coefficient. Colors from blue to red represent lower and higher strength of correlation.}
\end{figure}

\begin{figure*}
\centering
\includegraphics[width=0.9\textwidth]{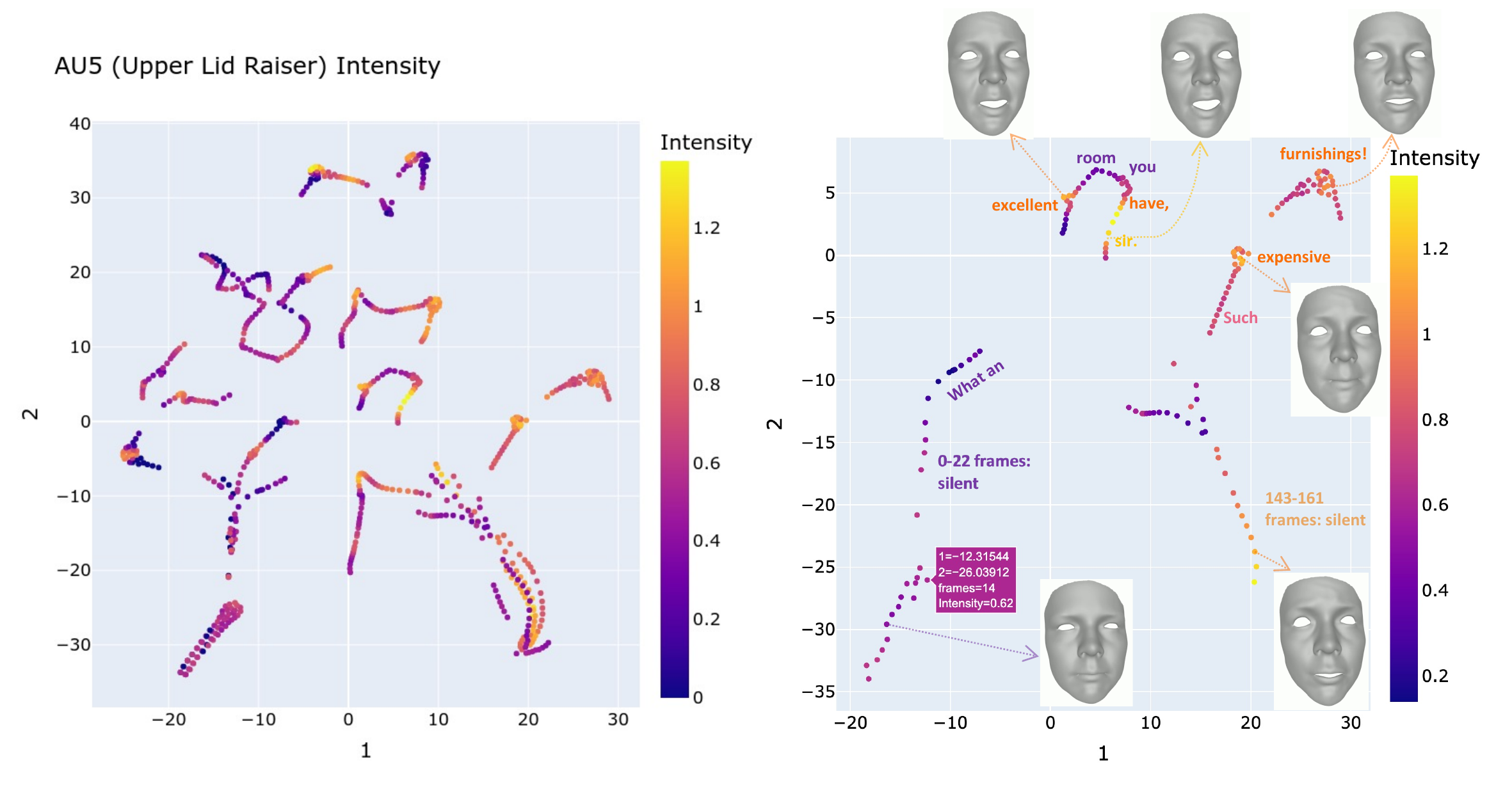}
\caption{\label{fig:visualization}
t-SNE visualization of the text embedding. The data points represent the projected text features. The color of each point corresponds to the AU5 intensity predicted from the corresponding rendered video frame. Left: the projected text features from all testing sentences of a subject. Right: the projected text features from one testing sentence of the same subject.
}
\end{figure*}

\subsection{Ablation Study}

\subsubsection{Ablation on the input modalities.}

To explore the importance of different input modalities, we carry out an ablation experiment to compare the performance among unimodal and bimodal models. From Table~\ref{tb:ablation}, the bimodal model ``Audio+Text'' performs best, outperforming the unimodal models ``Audio Only'' and ``Text Only''. It can be observed that using text alone has relatively large errors for the lower face AUs, whereas using audio alone degrade the performance for the upper face AUs. This suggests that the text information is beneficial in learning the variations in the upper face expressions and the audio information has a more strong correlation with the mouth movements. With both the audio and text modalities incorporated, the bimodal model leads to the best performance for the upper and lower face AUs. This indicates that audio and text modalities are complementary to each other. We also provide qualitative examples
for modality ablation in our supplementary video. From the videos, we find that using the text alone generates expressive muscle movements in the upper face part, e.g., opening the eyes wide when pronouncing the word ``thank''.

\subsubsection{Ablation on the fusion methods.}
Additionally, we conduct an experiment to compare the performance of the bimodal model using different fusion methods. As presented in Table~\ref{tb:ablation}, the tensor fusion strategy brings improvements compared to the simple concatenation method. This implies that the tensor fusion layer can better combine the audio and text cues. The 2-D Cartesian space might capture more variations in the speech, and hence allow a more expressive space for modeling the facial expression.  Besides, we conduct qualitative comparisons between the two fusion strategies to show the visual differences. As shown in the supplementary video, we observe that the results generated by the model with the tensor fusion strategy have better temporal stability in the mouth region.

\subsection{Analysis of the Text Embedding}
Figure~\ref{fig:visualization} shows a visualization of the extracted text features from the final output of the text encoder. We project the text features of all the testing sentences of a testing subject to 2-dimensional space using t-SNE. We can seen that the text features of the words within the same sentence are usually clustered. To better understand the text embeddings, we zoom in on the data points of one sentence (``What an excellent room you have, sir. Such expensive furnishings!'') of the same subject, as shown in the right side of Figure~\ref{fig:visualization}. Given the known duration time of each word within the sentence, we put the words close to the corresponding data point. We also visualize the corresponding generated facial expression to uncover how the facial expression changes when the word is changed. Colors from purple to yellow denote lower and higher AU5 (Upper Lid Raiser) intensity. In addition to the clusters, we observe that the AU5 intensity is higher for some words such as ``excellent''. According to the relationship between AUs and emotions~\cite{friesen1983emfacs}, AU5 is correlated with surprise, which also implicitly reflects the emotional state of the speaker. As can be noticed, there are some isolated data points in the lower part of the figure, meaning that those frames do not cover a word.

Additionally, we analyze which regions of the generated face mesh are influenced by the audio and text modalities. Specifically, after extracting audio and text features from their encoders, we calculate the absolute value of the Pearson correlation coefficient between the specific modality features and each predicted vertex offset (in total 23370 vertices). The greater the absolute value is, the stronger the relationship is. As shown in Figure~\ref{fig:correlation}, color codes are for strength of correlation. The Pearson analysis suggests that the audio modality has a strong association with vertices in the mouth region, and is moderately correlated with vertices in the cheek region. In contrast, the text modality shows a strong relationship with vertices in the upper face region, and also influences vertices in the cheek and upper lip regions.

\section{Conclusion, Limitations, and Future Work}
In this work, we present a novel approach for expressive speech-driven 3D facial animation based on audio and text. We pioneer the use of the pre-trained language models and feel that the contextual text embeddings offer great promise to this field for producing more expressive facial motions. Our method exhibits better performance against previous approaches, which we show through quantitative and qualitative analysis and perceptual user studies. However, the proposed model also presents limitations and sometimes fails to model the lip closure well when encountering certain plosives. On the other hand, due to the limitations of the current data, our model lacks some additional motions such as eye gaze and head movement. Future work may include studying the correlations between the speech signal and other subtle motions to achieve a higher level of conversational realism.

\newpage

\bibliography{aaai22}

\begin{thebibliography}{32}
\providecommand{\natexlab}[1]{#1}

\bibitem[{Baltru{\v{s}}aitis, Robinson, and
  Morency(2016)}]{baltruvsaitis2016openface}
Baltru{\v{s}}aitis, T.; Robinson, P.; and Morency, L.-P. 2016.
\newblock Openface: an open source facial behavior analysis toolkit.
\newblock In \emph{2016 IEEE Winter Conference on Applications of Computer
  Vision (WACV)}, 1--10. IEEE.

\bibitem[{Cao et~al.(2005)Cao, Tien, Faloutsos, and Pighin}]{cao2005expressive}
Cao, Y.; Tien, W.~C.; Faloutsos, P.; and Pighin, F. 2005.
\newblock Expressive speech-driven facial animation.
\newblock \emph{ACM Transactions on Graphics (TOG)}, 24(4): 1283--1302.

\bibitem[{Cudeiro et~al.(2019)Cudeiro, Bolkart, Laidlaw, Ranjan, and
  Black}]{cudeiro2019capture}
Cudeiro, D.; Bolkart, T.; Laidlaw, C.; Ranjan, A.; and Black, M.~J. 2019.
\newblock Capture, learning, and synthesis of 3D speaking styles.
\newblock In \emph{Proceedings of the IEEE/CVF Conference on Computer Vision
  and Pattern Recognition}, 10101--10111.

\bibitem[{Devlin et~al.(2018)Devlin, Chang, Lee, and
  Toutanova}]{devlin2018bert}
Devlin, J.; Chang, M.-W.; Lee, K.; and Toutanova, K. 2018.
\newblock Bert: Pre-training of deep bidirectional transformers for language
  understanding.
\newblock \emph{arXiv preprint arXiv:1810.04805}.

\bibitem[{Eckman and Friesen(1978)}]{eckman1978facial}
Eckman, P.; and Friesen, W. 1978.
\newblock Facial action coding system: A technique for the measurement of
  facial movement.
\newblock \emph{Consulting Psychologists Press}.

\bibitem[{Edwards et~al.(2016)Edwards, Landreth, Fiume, and
  Singh}]{edwards2016jali}
Edwards, P.; Landreth, C.; Fiume, E.; and Singh, K. 2016.
\newblock JALI: an animator-centric viseme model for expressive lip
  synchronization.
\newblock \emph{ACM Transactions on graphics (TOG)}, 35(4): 1--11.

\bibitem[{Fanelli et~al.(2010)Fanelli, Gall, Romsdorfer, Weise, and
  Van~Gool}]{fanelli2010}
Fanelli, G.; Gall, J.; Romsdorfer, H.; Weise, T.; and Van~Gool, L. 2010.
\newblock A 3-d audio-visual corpus of affective communication.
\newblock \emph{IEEE Transactions on Multimedia}, 12(6): 591--598.

\bibitem[{Fisher(1968)}]{fisher1968confusions}
Fisher, C.~G. 1968.
\newblock Confusions among visually perceived consonants.
\newblock \emph{Journal of speech and hearing research}, 11(4): 796--804.

\bibitem[{Friesen, Ekman et~al.(1983)}]{friesen1983emfacs}
Friesen, W.~V.; Ekman, P.; et~al. 1983.
\newblock EMFACS-7: Emotional facial action coding system.
\newblock \emph{Unpublished manuscript, University of California at San
  Francisco}, 2(36): 1.

\bibitem[{Hannun et~al.(2014)Hannun, Case, Casper, Catanzaro, Diamos, Elsen,
  Prenger, Satheesh, Sengupta, Coates et~al.}]{hannun2014deep}
Hannun, A.; Case, C.; Casper, J.; Catanzaro, B.; Diamos, G.; Elsen, E.;
  Prenger, R.; Satheesh, S.; Sengupta, S.; Coates, A.; et~al. 2014.
\newblock Deep speech: Scaling up end-to-end speech recognition.
\newblock \emph{arXiv preprint arXiv:1412.5567}.

\bibitem[{Karras et~al.(2017)Karras, Aila, Laine, Herva, and
  Lehtinen}]{karras2017audio}
Karras, T.; Aila, T.; Laine, S.; Herva, A.; and Lehtinen, J. 2017.
\newblock Audio-driven facial animation by joint end-to-end learning of pose
  and emotion.
\newblock \emph{ACM Transactions on Graphics (TOG)}, 36(4): 1--12.

\bibitem[{Kucherenko et~al.(2020)Kucherenko, Jonell, van Waveren, Henter,
  Alexandersson, Leite, and Kjellstr{\"o}m}]{kucherenko2020gesticulator}
Kucherenko, T.; Jonell, P.; van Waveren, S.; Henter, G.~E.; Alexandersson, S.;
  Leite, I.; and Kjellstr{\"o}m, H. 2020.
\newblock Gesticulator: A framework for semantically-aware speech-driven
  gesture generation.
\newblock In \emph{Proceedings of the 2020 International Conference on
  Multimodal Interaction}, 242--250.

\bibitem[{Massaro et~al.(2012)Massaro, Cohen, Tabain, Beskow, and
  Clark}]{massaro2012}
Massaro, D.; Cohen, M.; Tabain, M.; Beskow, J.; and Clark, R. 2012.
\newblock Animated speech: research progress and applications.
\newblock \emph{Audiovisual Speech Processing}, 309–345.

\bibitem[{McFee et~al.(2015)McFee, Raffel, Liang, Ellis, McVicar, Battenberg,
  and Nieto}]{mcfee2015librosa}
McFee, B.; Raffel, C.; Liang, D.; Ellis, D.~P.; McVicar, M.; Battenberg, E.;
  and Nieto, O. 2015.
\newblock librosa: Audio and music signal analysis in python.
\newblock In \emph{Proceedings of the 14th python in science conference},
  volume~8, 18--25. Citeseer.

\bibitem[{Ochshorn and Hawkins(2017)}]{ochshorn2017gentle}
Ochshorn, R.; and Hawkins, M. 2017.
\newblock Gentle forced aligner.
\newblock \emph{github. com/lowerquality/gentle}.

\bibitem[{Oord et~al.(2016)Oord, Dieleman, Zen, Simonyan, Vinyals, Graves,
  Kalchbrenner, Senior, and Kavukcuoglu}]{oord2016wavenet}
Oord, A. v.~d.; Dieleman, S.; Zen, H.; Simonyan, K.; Vinyals, O.; Graves, A.;
  Kalchbrenner, N.; Senior, A.; and Kavukcuoglu, K. 2016.
\newblock Wavenet: A generative model for raw audio.
\newblock \emph{arXiv preprint arXiv:1609.03499}.

\bibitem[{Pham, Cheung, and Pavlovic(2017)}]{pham2017speech}
Pham, H.~X.; Cheung, S.; and Pavlovic, V. 2017.
\newblock Speech-driven 3D facial animation with implicit emotional awareness:
  a deep learning approach.
\newblock In \emph{Proceedings of the IEEE Conference on Computer Vision and
  Pattern Recognition Workshops}, 80--88.

\bibitem[{Poria et~al.(2016)Poria, Chaturvedi, Cambria, and
  Hussain}]{poria2016convolutional}
Poria, S.; Chaturvedi, I.; Cambria, E.; and Hussain, A. 2016.
\newblock Convolutional MKL based multimodal emotion recognition and sentiment
  analysis.
\newblock In \emph{2016 IEEE 16th international conference on data mining
  (ICDM)}, 439--448. IEEE.

\bibitem[{Radford et~al.(2019)Radford, Wu, Child, Luan, Amodei, Sutskever
  et~al.}]{radford2019language}
Radford, A.; Wu, J.; Child, R.; Luan, D.; Amodei, D.; Sutskever, I.; et~al.
  2019.
\newblock Language models are unsupervised multitask learners.
\newblock \emph{OpenAI blog}, 1(8): 9.

\bibitem[{Ramesh et~al.(2021)Ramesh, Pavlov, Goh, Gray, Voss, Radford, Chen,
  and Sutskever}]{ramesh2021zero}
Ramesh, A.; Pavlov, M.; Goh, G.; Gray, S.; Voss, C.; Radford, A.; Chen, M.; and
  Sutskever, I. 2021.
\newblock Zero-shot text-to-image generation.
\newblock \emph{arXiv preprint arXiv:2102.12092}.

\bibitem[{Richard et~al.(2021)Richard, Zollhoefer, Wen, De~la Torre, and
  Sheikh}]{richard2021meshtalk}
Richard, A.; Zollhoefer, M.; Wen, Y.; De~la Torre, F.; and Sheikh, Y. 2021.
\newblock MeshTalk: 3D Face Animation from Speech using Cross-Modality
  Disentanglement.
\newblock \emph{arXiv preprint arXiv:2104.08223}.

\bibitem[{Sjerps et~al.(2019)Sjerps, Fox, Johnson, and
  Chang}]{sjerps2019speaker}
Sjerps, M.~J.; Fox, N.~P.; Johnson, K.; and Chang, E.~F. 2019.
\newblock Speaker-normalized sound representations in the human auditory
  cortex.
\newblock \emph{Nature communications}, 10(1): 1--9.

\bibitem[{Suwajanakorn, Seitz, and
  Kemelmacher-Shlizerman(2017)}]{suwajanakorn2017synthesizing}
Suwajanakorn, S.; Seitz, S.~M.; and Kemelmacher-Shlizerman, I. 2017.
\newblock Synthesizing obama: learning lip sync from audio.
\newblock \emph{ACM Transactions on Graphics (ToG)}, 36(4): 1--13.

\bibitem[{Taylor et~al.(2017)Taylor, Kim, Yue, Mahler, Krahe, Rodriguez,
  Hodgins, and Matthews}]{taylor2017deep}
Taylor, S.; Kim, T.; Yue, Y.; Mahler, M.; Krahe, J.; Rodriguez, A.~G.; Hodgins,
  J.; and Matthews, I. 2017.
\newblock A deep learning approach for generalized speech animation.
\newblock \emph{ACM Transactions on Graphics (TOG)}, 36(4): 1--11.

\bibitem[{Taylor et~al.(2012)Taylor, Mahler, Theobald, and
  Matthews}]{taylor2012dynamic}
Taylor, S.~L.; Mahler, M.; Theobald, B.-J.; and Matthews, I. 2012.
\newblock Dynamic units of visual speech.
\newblock In \emph{Proceedings of the 11th ACM SIGGRAPH/Eurographics conference
  on Computer Animation}, 275--284.

\bibitem[{Vaswani et~al.(2017)Vaswani, Shazeer, Parmar, Uszkoreit, Jones,
  Gomez, Kaiser, and Polosukhin}]{vaswani2017attention}
Vaswani, A.; Shazeer, N.; Parmar, N.; Uszkoreit, J.; Jones, L.; Gomez, A.~N.;
  Kaiser, {\L}.; and Polosukhin, I. 2017.
\newblock Attention is all you need.
\newblock In \emph{Advances in Neural Information Processing Systems},
  5998--6008.

\bibitem[{Wang, Fan, and Xia(2021)}]{wang20213d}
Wang, Q.; Fan, Z.; and Xia, S. 2021.
\newblock 3D-TalkEmo: Learning to Synthesize 3D Emotional Talking Head.
\newblock \emph{arXiv preprint arXiv:2104.12051}.

\bibitem[{Xu et~al.(2013)Xu, Feng, Marsella, and Shapiro}]{xu2013practical}
Xu, Y.; Feng, A.~W.; Marsella, S.; and Shapiro, A. 2013.
\newblock A practical and configurable lip sync method for games.
\newblock In \emph{Proceedings of Motion on Games}, 131--140.

\bibitem[{Yoon et~al.(2020)Yoon, Cha, Lee, Jang, Lee, Kim, and
  Lee}]{yoon2020speech}
Yoon, Y.; Cha, B.; Lee, J.-H.; Jang, M.; Lee, J.; Kim, J.; and Lee, G. 2020.
\newblock Speech gesture generation from the trimodal context of text, audio,
  and speaker identity.
\newblock \emph{ACM Transactions on Graphics (TOG)}, 39(6): 1--16.

\bibitem[{Yu and Koltun(2015)}]{yu2015multi}
Yu, F.; and Koltun, V. 2015.
\newblock Multi-scale context aggregation by dilated convolutions.
\newblock \emph{arXiv preprint arXiv:1511.07122}.

\bibitem[{Zadeh et~al.(2017)Zadeh, Chen, Poria, Cambria, and
  Morency}]{zadeh2017tensor}
Zadeh, A.; Chen, M.; Poria, S.; Cambria, E.; and Morency, L.-P. 2017.
\newblock Tensor fusion network for multimodal sentiment analysis.
\newblock \emph{arXiv preprint arXiv:1707.07250}.

\bibitem[{Zhou et~al.(2018)Zhou, Xu, Landreth, Kalogerakis, Maji, and
  Singh}]{zhou2018visemenet}
Zhou, Y.; Xu, Z.; Landreth, C.; Kalogerakis, E.; Maji, S.; and Singh, K. 2018.
\newblock Visemenet: Audio-driven animator-centric speech animation.
\newblock \emph{ACM Transactions on Graphics (TOG)}, 37(4): 1--10.

\end{thebibliography}

\end{document}